\documentclass[11pt,a4paper]{article}
\usepackage{times,latexsym}
\usepackage{url}
\usepackage[T1]{fontenc}
\usepackage{hyperref}
\usepackage{subfig}
\usepackage{mwe}
\usepackage{graphicx}
\usepackage{multirow}
\usepackage{array}
\usepackage{booktabs}
\usepackage{float} 
\usepackage{amssymb}
\usepackage{todonotes}
\usepackage{xcolor}
\usepackage{multirow,multicol}
\usepackage{times,latexsym}
\usepackage{url}
\usepackage[T1]{fontenc}
\usepackage{bm}
\usepackage{booktabs} 
\usepackage{multirow,multicol}
\usepackage{colortbl}
\usepackage{kotex}

    \usepackage[acceptedWithA]{tacl2018v2}
\usepackage[]{tacl2018v2}

\usepackage{xspace,mfirstuc,tabulary}

\newif\iftaclinstructions
\taclinstructionsfalse 
\iftaclinstructions
\renewcommand{\confidential}{}
\renewcommand{\anonsubtext}{(No author info supplied here, for consistency with
TACL-submission anonymization requirements)}
\newcommand{\instr}
\fi

\iftaclpubformat 

\else

\fi

\newcommand*\samethanks[1][\value{footnote}]{\footnotemark[#1]}

\title{Towards the Next 1000 Languages in Multilingual Machine Translation:\\ Exploring the Synergy Between Supervised and Self-Supervised Learning}

\author{
 Aditya Siddhant\thanks{*Equal contribution. Correspondence to \tt{\{adisid,ankurbpn,orhanf\}@google.com}} 
, Ankur Bapna\samethanks ~, Orhan Firat, Yuan Cao, Mia Xu Chen, Isaac Caswell, Xavier Garcia\\
 Google Research \\
}

\date{}

\begin{document}
\maketitle
\begin{abstract}

Achieving universal translation between all human language pairs is the holy-grail of machine translation (MT) research. While recent progress in massively multilingual MT is one step closer to reaching this goal, it is becoming evident that extending a multilingual MT system simply by training on more parallel data is unscalable, since the availability of labeled data for low-resource and non-English-centric language pairs is forbiddingly limited. To this end, we present a pragmatic approach towards building a multilingual MT model that covers hundreds of languages, using a mixture of supervised and self-supervised objectives, depending on the data availability for different language pairs. We demonstrate that the synergy between these two training paradigms enables the model to produce high-quality translations in the zero-resource setting, even surpassing supervised translation quality for low- and mid-resource languages. We conduct a wide array of experiments to understand the effect of the degree of multilingual supervision, domain mismatches and amounts of parallel and monolingual data on the quality of our self-supervised multilingual models. To demonstrate the scalability of the approach, we train models with over 200 languages and demonstrate high performance on zero-resource translation on several previously under-studied languages. We hope our findings will serve as a stepping stone towards enabling translation for the next thousand languages.
\end{abstract}

\section{Introduction}

\begin{figure}[t!]
\begin{center}
\includegraphics[scale=0.3]{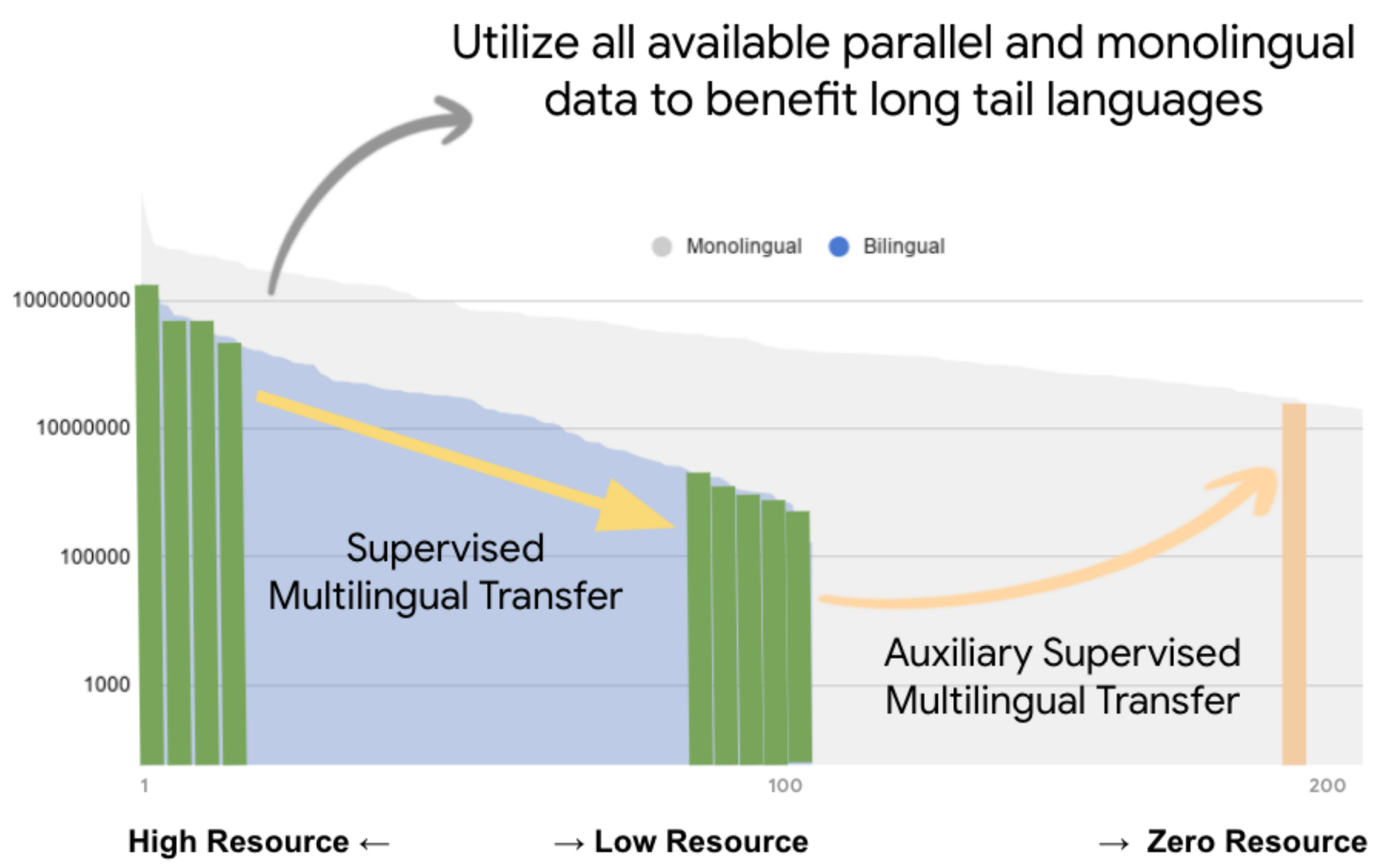}
\caption{Plot demonstrating the relative sizes of parallel and monolingual examples in our web-mined data, and the direction of transfer from supervised high and low resource language pairs to benefit zero-resource languages.}
\label{fig:data}
\end{center}
\end{figure}

The success of Neural Machine Translation (NMT) has relied on the availability of large amounts of parallel data, often mined from the web \cite{wu2016google}. However, the amount of readily available bilingual data falls off from billions of sentence pairs for high resource languages like Spanish and German, to a few tens of thousands for Sindhi and Hawaiian \cite{arivazhagan2019massively}. The limited availability of bilingual data has spurred research into massively multilingual MT \cite{dong-etal-2015-multi,firat16,firat-etal-2016-zero,ha2016universal,johnson2017google,aharoni-etal-2019-massively,arivazhagan2019massively}; the large amounts of supervision available for high resource languages has been harnessed to boost quality for languages with relatively limited amounts of parallel data \cite{zoph16,DBLP:journals/corr/abs-1708-09803,neubig2018rapid}. However, beyond the highest resource 100 languages, bilingual data is a scarce resource often limited only to narrow domain religious texts.

Another thread of research has pursued learning MT models directly from monolingual data \cite{artetxe2017unsupervised,lample2018unsupervised,lample2018phrase,song2019mass,lewis2019bart}. While unsupervised MT approaches have recently started getting close to the quality of fully supervised systems, these approaches are typically brittle, and rely on the availability of large amounts of domain matched monolingual datasets across the source and target languages \cite{marchisio2020does,kim2020and}; a luxury not available for real-world low resource languages.

In this paper we study what it takes to extend the language coverage of massively multilingual models from a practical point of view. We take a pragmatic stance: (i) we do not constrain the approaches to be purely unsupervised, and attempt to benefit from all available parallel data at hand with direct or indirect supervision, and (ii) we devise practical methods that make use of monolingual and parallel data simultaneously without resorting to complex techniques or tedious pipelining of data augmentation techniques. Our proposed approach scales elegantly and efficiently, which we showcase by extending massively multilingual MT models to beyond 200 languages.

Our study aims at confluencing multilingual supervised and unsupervised MT research by demonstrating that multilingual models can mitigate the deficiencies of unsupervised MT as they cover a wide range of domains, languages, and scripts. Conversely, massively multilingual MT can benefit from unsupervised MT methods to better make use of monolingual data \cite{siddhant2020leveraging,garcia20,garcia2021harnessing}.  We analyze the effects of increasing the number of supervised languages, self-supervised languages, domain mismatch, and the amounts of available monolingual and parallel data on the performance of our multilingual, self-supervised models.\footnote{By supervised and self-supervised languages we refer to the type of data and objectives being utilized for a particular language in our MT model. If any parallel data is utilized for a given language, we define it to be supervised in the model. If only monolingual data is used we deem it self-supervised.} 
Our results indicate that training a massively multilingual model with large amounts of indirect supervision results in robust translation models capable of circumventing the weaknesses of traditional bilingual unsupervised MT models.

We apply our approach to a real world setting by training a 200 language MT model on a large web-scale dataset, using parallel data from only 100 languages. Our model demonstrates strong Many-to-English (xx $\rightarrow$en) translation quality on most of our evaluations, and promising English-to-Many (en $\rightarrow$xx) quality when enough monolingual data is available.

\paragraph{Terminology} In our pragmatic setting, a multilingual model has some language pairs for which we have parallel data available and some language pairs for which we rely only on monolingual data. If parallel data is available for a language pair, we refer to that setting as \textit{supervised} in this paper while the language pairs with only monolingual data are interchangeably referred to as \textit{self-supervised} or \textit{zero-resource} language pair. 

\section{Method}
We propose a co-training mechanism that combines supervised multilingual NMT with monolingual data and self-supervised learning. While several pre-training based approaches have been studied in the context of NMT \cite{dai2015semi,conneau2019cross,song2019mass}, we proceed with Masked Sequence-to-Sequence (MASS) \cite{song2019mass} given its success on unsupervised and low-resource NMT, and adapt it to the multilingual setting.

\subsection{Adapting MASS for multilingual models}
MASS adapts the masked de-noising objective \citep{devlin2019bert,raffel2019exploring} for sequence-to-sequence models by masking the input to the encoder and training the decoder to generate the masked portion of the input. To utilize this objective function for unsupervised NMT, \citet{song2019mass} enhance their model with additional improvements, including language embeddings, target language-specific attention context projections, shared target embeddings and softmax parameters and high variance uniform initialization for target attention projection matrices\footnote{Verified from open-source Github implementation.}.

We use the same set of hyper-parameters for self-supervised training as described in \cite{song2019mass}. However, while the success of MASS relies on the architectural \textit{modifications} described above, 
we find that our multilingual NMT experiments are stable even in the absence of these techniques, thanks to the smoothing effect of multilingual joint training \citep{siddhant2020leveraging}. We also forego separate source and target language embeddings in favour of prepending the source sentences with a \texttt{<2xx>} token \cite{johnson2017google}. We train our models simultaneously on supervised parallel data using the translation objective and on monolingual data using the MASS objective.

Note that in our formulation, the \texttt{<2xx>} token signals both the MASS task and the translation task -- in effect it conveys the message ``convert the input into a fluent, semantically equivalent sentence in the language `xx', regardless of the input form".

\subsection{Iterative Back Translation}

A model trained only on translation and MASS is already capable of generating high-quality xx$\rightarrow$en \ (to English) translations for all of the low-resource pairs we consider. Unfortunately, the model does not perform well on the en$\rightarrow$xx \, language pairs. Note that this phenomenon is ubiquitously observed in multilingual models \cite{firat16,johnson16,aharoni-etal-2019-massively}. We find that an additional training stage incorporating iterative back-translation alleviates performance bottlenecks in the en$\rightarrow$xx \ direction. The strong xx$\rightarrow$en translation provides the pseduo-parallel data for en$\rightarrow$xx direction. Please refer to Step 2 in \citet{garcia20} for more details about this algorithm.

\begin{table*}[t]
\centering
\begin{tabular}{l|cc|cc|cc|cc|cc}
\toprule
                   & \multicolumn{2}{c|}{En-Gu}    & \multicolumn{2}{c|}{En-Tr}     & \multicolumn{2}{c|}{En-Kk}     & \multicolumn{2}{c|}{En-Hi}    & \multicolumn{2}{c}{En-Ro}    \\
Para Size          & \multicolumn{2}{c|}{0.01m}    & \multicolumn{2}{c|}{0.2m}      & \multicolumn{2}{c|}{0.2m}      & \multicolumn{2}{c|}{0.3m}     & \multicolumn{2}{c}{0.6m}     \\
Mono Size          & \multicolumn{2}{c|}{4.6m}     & \multicolumn{2}{c|}{9.6m}      & \multicolumn{2}{c|}{13.8m}     & \multicolumn{2}{c|}{23.6m}    & \multicolumn{2}{c}{14.1m}    \\

            Direction       & $\leftarrow$ & $\rightarrow$ & $\leftarrow$  & $\rightarrow$ & $\leftarrow$  & $\rightarrow$ & $\leftarrow$ & $\rightarrow$ & $\leftarrow$ & $\rightarrow$ \\
                   \midrule
Multilingual  & 3.0                            & 0.4                            & \textbf{18.4} & \textbf{17.4} & 11.2                           & 2.3                            & 16.0                           & 11.6                           & 33.0                           & 24.3                           \\
Zero-resource & \textbf{14.8} & \textbf{11.7} & 15.8                           & 16.4                           & \textbf{12.8} & \textbf{9.3}  & \textbf{17.2} & \textbf{11.8} & \textbf{36.8} & \textbf{25.4} \\
\midrule
\midrule
                   & \multicolumn{2}{c|}{En-Lt}    & \multicolumn{2}{c|}{En-Lv}     & \multicolumn{2}{c|}{En-Et}     & \multicolumn{2}{c|}{En-De}    & \multicolumn{2}{c}{En-Fi}    \\
Para Size          & \multicolumn{2}{c|}{0.6m}     & \multicolumn{2}{c|}{0.6m}      & \multicolumn{2}{c|}{2.1m}      & \multicolumn{2}{c|}{4.5m}     & \multicolumn{2}{c}{6.5m}     \\
Mono Size          & \multicolumn{2}{c|}{106.1m}   & \multicolumn{2}{c|}{10.2m}     & \multicolumn{2}{c|}{51.6m}     & \multicolumn{2}{c|}{275m}     & \multicolumn{2}{c}{18.8m}    \\
      Direction             & $\leftarrow$ & $\rightarrow$ & $\leftarrow$  & $\rightarrow$ & $\leftarrow$  & $\rightarrow$ & $\leftarrow$ & $\rightarrow$ & $\leftarrow$ & $\rightarrow$ \\
      \midrule
Multilingual  & 24.3                           & \textbf{12.4} & 17.6                           & 15.5                           & \textbf{24.2} & 17.9                           & 33.2                           & 26.4                           & \textbf{25.2} & \textbf{19.2} \\
Zero-resource & \textbf{24.4} & 8.9                            & \textbf{20.9} & \textbf{15.8} & 22.1                           & \textbf{18.9} & \textbf{35.9} & \textbf{26.7} & 23.8                           & 17.8                           \\
\midrule
\midrule
                   & \multicolumn{2}{c|}{En-Es}    & \multicolumn{2}{c|}{En-Zh}     & \multicolumn{2}{c|}{En-Ru}     & \multicolumn{2}{c|}{En-Fr}    & \multicolumn{2}{c}{En-Cs}    \\
Para Size          & \multicolumn{2}{c|}{15.1m}    & \multicolumn{2}{c|}{25.9m}     & \multicolumn{2}{c|}{38.4m}     & \multicolumn{2}{c|}{40.4m}    & \multicolumn{2}{c}{64.3m}    \\
Mono Size          & \multicolumn{2}{c|}{43.8m}    & \multicolumn{2}{c|}{72.1m}      & \multicolumn{2}{c|}{80.1m}     & \multicolumn{2}{c|}{160.9m}   & \multicolumn{2}{c}{72.1m}    \\
         Direction          & $\leftarrow$ & $\rightarrow$ & $\leftarrow$  & $\rightarrow$ & $\leftarrow$  & $\rightarrow$ & $\leftarrow$ & $\rightarrow$ & $\leftarrow$ & $\rightarrow$ \\
         \midrule

Multilingual  & 28.9                           & \textbf{30.1} & \textbf{17.7} & \textbf{24.6} & \textbf{33.8} & 21.8                           & 34.9                           & \textbf{37.5} & \textbf{28.4} & 18.9                           \\
Zero-resource & \textbf{29.1} & 28.5                           & 16.9                           & 21.2                           & 30.6                           & \textbf{25.5} & \textbf{36.1} & 33.2                           & 28.3                           & \textbf{22.1} \\
\bottomrule
\end{tabular}
\caption{Results of our zero-resource models, which do not use parallel data for the language pair they are evaluated on, compared to the multilingual baseline which is trained on all language pairs using all parallel data available in the WMT corpus. The languages have been sorted from low to high resource depending on the availability of the parallel data.} \label{tab:main}
\end{table*}

\section{A Study with WMT data}
\label{sec:wmt}

We first conduct an extensive study of our approach with a WMT dataset. We train 15 different multilingual models on WMT parallel and monolingual data. In each of these models, we leave out parallel data for one of the languages. We then evaluate this model on the language for which only monolingual data was used, in order to test the zero-resource performance. Please note that we leave out the parallel data to simulate the pragmatic setting where we do not have parallel data for all language pairs. 
All our models are simultaneously trained on both en$\rightarrow$xx and xx$\rightarrow$en translation tasks.

\subsection{Experimental Setup}

\paragraph{Dataset}
We use the parallel and monolingual training data provided with the WMT corpus, for 15 languages to and from English. The amount of parallel data available ranges from more than 60 million sentence pairs, as in En-Cs, to roughly 10k sentence pairs for En-Gu. We also collect additional monolingual data from WMT news-crawl, news-commentary, common-crawl, europarl-v9, news-discussions and wikidump datasets in all 16 languages including English.\footnote{Followed the versions recommended by WMT'19 shared task, as in http://statmt.org/wmt19/translation-task.html} The amount of monolingual data varies from 2 million sentences in Zh to 270 million in De. The amount of the parallel and monolingual data used for experiments in the paper is depicted in Table~\ref{tab:main}.

\paragraph{Data Sampling}
Given the data imbalance across languages in our datasets, we use a temperature-based data balancing strategy to over-sample low-resource languages in our multilingual models \citep{arivazhagan2019massively}. We use a temperature of $T=5$ to balance our parallel training data. When applicable, we sample monolingual data uniformly across languages since this distribution is not as skewed. For experiments that use both monolingual and parallel data, we mix the two sources at an equal ratio (50\% monolingual data with self-supervision and 50\% parallel data). 

\paragraph{Hyperparameters}
All experiments are performed with the Transformer architecture \citep{vaswani2017attention} using the open-source Tensorflow-Lingvo implementation \citep{lingvo}. Specifically, we use the Transformer Big model containing 375M parameters (6 layers, 16 heads, 8192 hidden dimension) \citep{chen2018best} and a shared source-target SentencePiece model (SPM)\footnote{https://github.com/google/sentencepiece} \cite{kudo2018sentencepiece}. We use a vocabulary size of 64k for all the models. Different SPMs are trained depending on the set of languages supported by the model. Both stages of training use the Adafactor optimizer \cite{shazeer2018adafactor} and a batch size of 1024. The first stage of training runs for 500k steps and we use validation set performance for checkpoint picking. Iterative back translation runs for another 100k steps.

\paragraph{Evaluation}
We evaluate the performance of the models using
SacreBLEU on standard WMT validation and test sets \cite{post-2018-call}. The year of dev and test sets used is reported in Appendix Table \ref{appendix:tab:paralleldata}.

\begin{table*}[t]
\centering
\begin{tabular}{cc|cc|cc|cc|cc}
\toprule
       &        & \multicolumn{2}{c|}{En-Fr}    & \multicolumn{2}{c|}{En-Ro}    & \multicolumn{2}{c|}{En-Hi}    & \multicolumn{2}{c}{En-Tr}    \\
       \midrule
\#Para & \#Mono & $\leftarrow$ & $\rightarrow$ & $\leftarrow$ & $\rightarrow$ & $\leftarrow$ & $\rightarrow$ & $\leftarrow$ & $\rightarrow$ \\
\midrule 
1      & 1      & 21.9         & 21.3          & 24.9         & 14.9          & 7.2          & 6.2           & 8.7          & 7.3           \\
4      & 1      & 32.9         & 30.8          & 32.9         & 22.8          & 11.8         & 9.9           & 11.9         & 12.6          \\
8      & 1      & 34.2         & 32.3          & 35.3         & 24.1          & 14.9         & 10.8          & 14.9         & 13.1          \\
14     & 1      & \textbf{36.1 }        & \textbf{33.2 }         & \textbf{36.8}         & \textbf{25.4}          & \textbf{17.2}         & \textbf{11.3}          & \textbf{15.8}         & \textbf{15.4}   \\
\bottomrule
\end{tabular}
\caption{Impact of number of languages with parallel data in the model. Keeping the number of parallel training examples the same, we increase the number of the parallel directions in the model from top to bottom. All numbers are zero-resource translation.} \label{tab:numparalleldirections}
\end{table*}

\begin{table*}[]
\centering
\begin{tabular}{cc|cc|cc|cc|cc|cc}
\toprule
       &        & \multicolumn{2}{c|}{En-Ro}                                            & \multicolumn{2}{c|}{En-Hi}    & \multicolumn{2}{c|}{En-Kk}    & \multicolumn{2}{c|}{En-Tr}    & \multicolumn{2}{c}{En-Gu}    \\
       \midrule
\#Para & \#Mono & $\leftarrow$ & $\rightarrow$ & $\leftarrow$ & $\rightarrow$ & $\leftarrow$ & $\rightarrow$ & $\leftarrow$ & $\rightarrow$ & $\leftarrow$ & $\rightarrow$ \\
\midrule
14     & 1      & \textbf{36.8}         & \textbf{25.4}          & \textbf{17.2}         & \textbf{11.3}          & \textbf{12.8}         & \textbf{9.3}           & \textbf{15.8}         & \textbf{15.8}          & \textbf{14.7}          & \textbf{14.8}          \\
10     & 5      & 31.9         & 23.3          & 15.8         & 9.9           & 12.1         & 8.1           & 14.2         & 12.9          & 12.3         & 11.9          \\
5      & 10     & 28.3         & 18.2          & 14.1         & 6.4           & 9.7          & 5.2           & 13.1         & 8.9           & 11.4         & 7.9           \\
1      & 14     & 18.7         & 12.3          & 4.9          & 4.4           & 3.7          & 4.1           & 7.8          & 5.1           & 5.4         & 4.1           \\
\bottomrule
\end{tabular}
\caption{Effects of adding multiple unsupervised languages to the same multilingual model. The number of unsupervised languages increases from top to bottom, which shows the edge case of having parallel data for only one language pair and only monolingual data for the other 14 languages. All numbers are zero-resource translation.}\label{tab:numunsuperviseddirections}
\end{table*}

\subsection{Results}
Our multilingual model in zero-resource setting is able to match the performance of fully supervised  multilingual baselines (of same capacity) in high resource languages and improve over it by significant margins for medium/low resource languages, despite not using any parallel data for the evaluated language pair. For the lowest-resource languages like
Kazakh (kk) and Gujarati (gu), we
can see from BLEU scores in \ref{tab:main} that multilingual NMT alone is not sufficient to reach high translation quality. The addition
of monolingual data has a large positive impact
on very low resource languages, significantly improving quality over the supervised multilingual
model. These improvements range are more pronounced in xx $\rightarrow$ en direction. Even for high resource languages, the method can achieve similar translation quality by leaving out parallel data entirely (for language under evaluation) and throwing in 3-4 times monolingual examples, which would be easier to obtain.

These results suggest that for a language pair without any parallel data, it is possible to get good quality translation (as measured by BLEU score) by relying on transfer from other supervised language pairs and self-supervision using monolingual data. One could think of this as monolingual data and self supervised objectives such as MASS helping the model learn the language and the supervised translation in other language pairs teaching the model how to translate by transfer learning. 

\begin{table*}[]
\centering
\begin{tabular}{cc|cc|cc|cc|cc}
\toprule
          &              & \multicolumn{2}{c|}{En-Ro}    & \multicolumn{2}{c|}{En-Gu}    & \multicolumn{2}{c|}{En-Kk}    & \multicolumn{2}{c}{En-Et}    \\
         \midrule
\#Similar & \#Dissimilar & $\leftarrow$ & $\rightarrow$ & $\leftarrow$ & $\rightarrow$ & $\leftarrow$ & $\rightarrow$ & $\leftarrow$ & $\rightarrow$ \\
\midrule
4         & 0            & \textbf{33.3}         & \textbf{23.2}          & \textbf{14.1}         & \textbf{13.9}          & \textbf{12.9}         & \textbf{9.1}           & \textbf{21.2}        & \textbf{17.3}          \\
2         & 2            & 29.9         & 20.9          & 13.9         & 12.1          & 9.9          & 7.9           & 18.2         & 14.9          \\
0         & 4            & 20.1         & 13.9          & 11.1         & 9.4           & 5.9          & 3.2           & 13.1         & 7.8           \\
\bottomrule
\end{tabular}
\caption{The choice of language pairs included in the multilingual model impacts the unsupervised translation in target language depending on the linguistic similarity between the languages. From top to bottom, we decrease the number of linguistically related parallel languages in the model.} \label{tab:similarity}
\end{table*}

\begin{table*}[]
\centering
\begin{tabular}{ll|cccc||ll|cccc}
\toprule
\multirow{2}{*}{\begin{tabular}[c]{@{}c@{}}Mono\\ Size\end{tabular}} & \multirow{2}{*}{\begin{tabular}[c]{@{}c@{}}Para\\ Size\end{tabular}} & \multicolumn{2}{c}{En-De}    & \multicolumn{2}{c||}{En-Hi}    & \multirow{2}{*}{\begin{tabular}[c]{@{}c@{}}Mono \\ Size\end{tabular}} & \multirow{2}{*}{\begin{tabular}[c]{@{}c@{}}Para\\ Size\end{tabular}} & \multicolumn{2}{c}{En-De}    & \multicolumn{2}{c}{En-Hi}    \\
                                                                     &                                                                      & $\leftarrow$ & $\rightarrow$ & $\leftarrow$ & $\rightarrow$ &                                                                       &                                                                      & $\leftarrow$ & $\rightarrow$ & $\leftarrow$ & $\rightarrow$ \\
\midrule
0.1m & 100m & 17.1         & 14.9          & 5.8          & 6.7           & 25m & 0.1m & 11.9         & 12.8          & 5.1          & 6.1           \\
1m   & 100m & 22.9         & 18.2          & 10.1         & 9.0           & 25m & 1m   & 18.6         & 17.0          & 9.2          & 8.5           \\
5m   & 100m & 26.4         & 20.8          & 13.0         & 10.1          & 25m & 10m  & 25.1         & 19.9          & 13.1         & 10.0          \\
25m  & 100m & \textbf{27.3}         & \textbf{22.1}          & \textbf{14.4}         & \textbf{10.8}          & 25m & 100m & \textbf{27.3}         & \textbf{22.1}          & \textbf{14.4}         & \textbf{10.8}          \\
\bottomrule
\end{tabular}
\caption{The impact of the amount of monolingual and parallel data in multilingual models on unsupervised performance in different settings. On the left, we increase the amount of monolingual data from top to bottom keeping parallel data constant while on the right we increase the amount of parallel data. All models are trained with supervised data in fr, es, cs, and ru.
} \label{tab:amount}
\end{table*}

\subsection{Analysis}
In the following segment, we conduct a systematic evaluation of the various criterion that affect the quality zero-resource/self-supervised languages within the model. Please note that the choice of language pairs we evaluate and train on varies for some of these experiments. We explain the choice of these languages in the appendix.

\subsubsection*{What happens if we add more languages with parallel data ?}

Adding more parallel directions while controlling for the number of training examples, i.e. increasing multilinguality, helps the model in all cases. This is intuitive, as having more languages with parallel data means that the unsupervised language learns something from each supervised pair. In Table \ref{tab:numparalleldirections}, we can see that as we add more languages with parallel data, performance of the unsupervised pair improves for all languages pairs. Another interesting takeaway from this experiment is that even though multilinguality always helps, a single high-resource supervised pair is enough to get unsupervised performance of 20 BLEU on some language pairs, demonstrating the effectiveness of this approach. Please note that these experiments were run individually for each target unsupervised language and all models trained for this experiment have only one unsupervised language. We control for the total amount of supervised data in order to distinguish the effect of additional multilinguality and additional parallel data from the same language.

\subsubsection*{What happens when we add multiple zero-resource languages in the same model ?}

Increasing the number of self-supervised languages in the model while decreasing the number of supervised languages diminishes performance. We find that there must be a certain number of supervised directions in the multilingual model for the approach to work. The edge case with using only one language with parallel data and monolingual data for 14 other languages fails to reach reasonable performance, and we see BLEU < 10 for most target language pairs. See Table \ref{tab:numunsuperviseddirections} for results of this simulation.

\begin{table*}[]
\centering
\begin{tabular}{l|ll|cc}
\toprule
\multicolumn{1}{c}{Mono} & \multicolumn{2}{c}{Para}                             & \multicolumn{2}{c}{En-Lt}    \\
\midrule
\multicolumn{1}{c}{}     & \multicolumn{1}{c}{Train} & \multicolumn{1}{c}{Eval} & $\rightarrow$ & $\leftarrow$ \\
\midrule
News Crawl                & News Crawl                 & News Crawl                & \textbf{18.4}          &   \textbf{8.6}           \\
Paracrawl                & News Crawl                 & News Crawl                & 17.4          &   8.4           \\
News Crawl                & CommonCrawl               & News Crawl                & 17.1          &    8.2          \\
News Crawl                & Wikipedia                 & News Crawl                & 17.7          &     7.9         \\
News Crawl                & Bible                    & News Crawl                & 7.9           &    4.1  \\
Paracrawl                & CommonCrawl               & News Crawl                & 14.8          &     7.1         \\

\bottomrule
\end{tabular} 
\caption{Impact of the domain of monolingual and parallel training data on performance. Each row is an experiment with a different combination of monolingual and parallel training data domains.} \label{tab:domain}
\end{table*}

\begin{table*}[]
\centering
\begin{tabular}{l|cccc}
\toprule
\multicolumn{1}{c}{Method}                                                 & \multicolumn{2}{c}{En-Ro}                   & \multicolumn{2}{c}{En-Lt}                   \\
\multicolumn{1}{c}{}                                                       & $\leftarrow$ & $\rightarrow$         & $\leftarrow$ & $\rightarrow$        \\
\midrule
Mono PT                                                                            & 6.8          & 7.3           & 4.8          & 2.1           \\
Mono PT $\rightarrow$ Para FT                                                      & 22.1         & 8.1           & 13.9         & 3.8           \\
Mono + Para CT                                                                     & 33.0         & 9.3           & 21.3         & 6.7           \\
Mono + Para CT $\rightarrow$ Target Lang IBT                                       & 35.2         & 24.1          & 22.0         & 8.1           \\
Mono + Para CT $\rightarrow$ Target Lang IBT + Mono FT                             & 35.6         & 24.4          & 22.1         & 8.1           \\
Mono + Para CT $\rightarrow$ Target Lang IBT + Para FT                             & \textbf{36.9}         & 25.1          & 24.0         & 8.8           \\
\multicolumn{1}{c|}{Mono + Para CT $\rightarrow$ Target Lang IBT + Mono + Para CT} & 36.8         & \textbf{25.4}          & \textbf{24.4}         & \textbf{8.9 }          \\
\bottomrule
\end{tabular}
\caption{Different ways of stacking the monolingual and parallel objectives during  training. Here, PT refers to pretraining, FT to finetuning, CT to co-training and IBT to iterative back translation.} \label{tab:stacking}
\end{table*}

\subsubsection*{How much do languages with related embeddings help ?}
\label{main:linguistic}

Intuitively, ``similar" languages should help more than unrelated languages. As a proxy for (dis)similarity, we chose languages whose embeddings on multiway parallel text were more similar or dissimilar based on \citet{kudugunta2019investigating} (See Figure 2 in their paper). We experimented with three settings: one where all parallel directions were similar to unsupervised target language, one where all were unrelated, and one middle-ground where two languages were related and two were unrelated. All results support the hypothesis that groupings of ``similar" languages improve transfer. Refer to Table \ref{tab:similarity} for the results of this experiment. We also observed that just having similar languages is not enough if all similar languages are low resource and have little parallel data. In that case having a high resource language is important for the model to reach high translation quality.

Please note that we use a definition of language similarity based on representations learnt within a massively multilingual MT model;  these sets of ``similar" languages are not necessarily similar from a linguistic perspective. For some languages (like Romanian) our definition of similarity does correspond with language families, for others (like Gujarati) that is not the case.  Details of the groupings may be found in Appendix \ref{appendix:linguistic}.

\subsubsection*{How much monolingual and parallel data do we need for these models to reach reasonable translation quality ?}
To study the impact of the amount of monolingual and parallel data, we conduct a set of experiments where we: (i) vary the amount of monolingual data from 100k to 25m for self-supervised languages (keeping the parallel data constant), and (ii) vary the amount of parallel data in supervised directions from 100k to 100m (keeping the monolingual data available for the self-supervised language constant). We find that the lack of parallel data is slightly more detrimental to performance compared to a lack of monolingual data. An interesting takeaway from this experiment is that with only 100k monolingual sentences, it it possible to reach a zero-resource BLEU of 17 on de-en translation if we have enough parallel data in other languages.

\subsubsection*{What happens if there is a domain mismatch in parallel and monolingual data sources ?} Table \ref{tab:domain} depicts the results of  experiments that analyze the impact of train and test domains of supervised and self-supervised directions. We first conduct an experiment where both monolingual and parallel training data are drawn from the same domain on which we evaluate the model: newscrawl. We then change the training data domain and find that while the model is more robust to domain mismatch as compared to completely unsupervised baselines, domain mismatch does hurt performance. In the final row of Table \ref{tab:domain}, we notice a significant drop in performance when both monolingual and parallel training data differ from the evaluation domain.

\subsubsection*{How to stack different stages of training ?} 

Finally, we compare different settings for training our proposed multilingual models. First we compare a two-stage process of pre-training with a self-supervised objective before finetuning with a supervised dataset \cite{lewis2019bart,song2019mass} against a single stage process where we train with both losses simultaneously. We find that the latter is significantly better, perhaps mitigating catastrophic forgetting \citep{McCloskey1989CatastrophicII}.
Subsequently, we compare which training data and objectives should be enabled in the second stage of the training along with iterative back translation. We find that while enabling monolingual and parallel finetuning on other languages is helpful, the benefit of doing so is not significant.

\section{Scaling to 200+ languages}

We next evaluate our approach in a more realistic setting, scaling up a single multilingual MT model to 200 languages on a web-scale dataset.

\begin{figure*}[h]
  \centering
  \subfloat[Any-to-English (xx $\rightarrow$ en) ]{\includegraphics[width=0.49\textwidth]{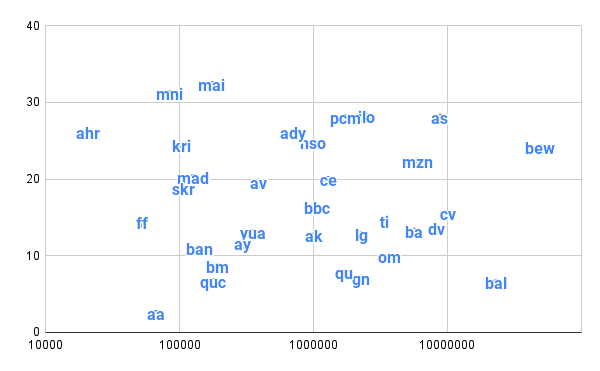} \label{fig:xxen}}
  \subfloat[English-to-Any (en $\rightarrow$ xx)]{\includegraphics[width=0.49\textwidth]{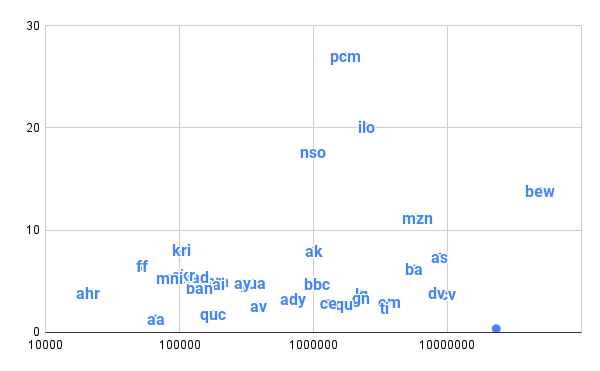} \label{fig:enxx}}
  \caption{Unsupervised/zero-resource BLEU on 30 new languages. The x-axis depicts the amount of monolingual data available for the language, while the y-axis depicts the BLEU score of the 1.6B parameter Transformer model after fine-tuning with online back-translation. The data point corresponding to each language is represented by its BCP-47 language code.}
  \label{fig:lang200}
\end{figure*}

\subsection{Experimental Setup}
\paragraph{Dataset}
Although the WMT data is high-quality, it is on the smaller side, and only exists for a relatively small subset of languages. To evaluate our approach in a realistic setting, we performed a deep, highly multilingual crawl of the web for monolingual and parallel data, using language predictions from an n-gram LangID model trained on trusted, in-house data. The parallel dataset covered 112 languages, with a median dataset size of 17M noisy sentence pairs per language.  For the monolingual dataset, where we expected to find more data, we covered a total of 206 languages, with a median of 4.4M sentences per language. For the 100 lowest-resource languages, we found that our data was very noisy and had a lot of over-triggering with higher-resource languages, and therefore additionally applied data filtering based on semi-supervised LangID and lists of highly indicative words in these languages (``TF-IIF"), following \citet{caswell2020language}. In order to balance precision and recall, we used tighter wordlist-filtering to produce a second, cleaner version of the monolingual dataset for these tail 100 languages, which averaged about half the size of the original dataset. The cleaner dataset was used for back-translation while the noisy unfiltered dataset was utilized for training with the MASS self-supervised objective.

\paragraph{Approach} Following the setup described in Section~\ref{sec:wmt}, we train a Transformer on the combined dataset. We increase the number of layers in the encoder and the decoder to 32 in order to reduce capacity limitations~\citep{arivazhagan2019massively,huang2019gpipe} for the increased number of languages and data, resulting in a model with 1.6B parameters. We use a SentencePiece vocabulary with 64k tokens covering all 206 languages to train this model. For all languages except English, the vocabulary is close to character-level.

The model is first pre-trained on the translation and MASS tasks for 250k steps with a batch size of 4M tokens per batch. Following this, the model is used to generate synthetic data for the zero-resource language pairs, using xx$\rightarrow$en translation on the cleaner version of the monolingual data. This data is filtered to remove copies and repetitions, and is used to fine-tune the model with back-translation and self-training~\citep{zhang-zong-2016-exploiting} tasks for 50k additional steps. We simultaneously train with online back-translation on all zero resource languages, and continue training the model with the parallel and monolingual datasets. We collected multi-way parallel evaluation sets for 30 of these 106 languages, attempting to prioritize linguistic and geographical diversity among this set of languages.

\subsection{Results}
The results of this experiment are depicted in Figure~\ref{fig:lang200}. We find that xx$\rightarrow$en and en$\rightarrow$xx translation quality exhibit different trends.

\begin{figure}
  \centering
  \includegraphics[width=0.49\textwidth]{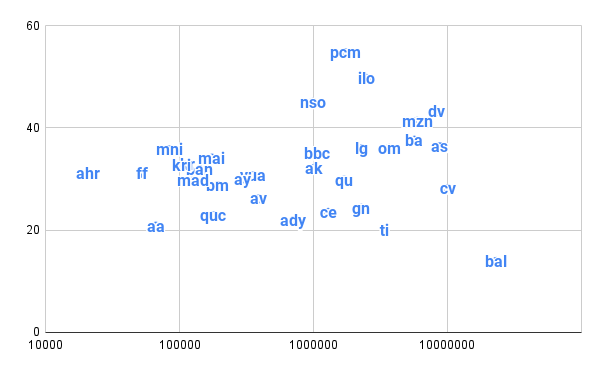}
  \caption{Unsupervised/zero-resource English-to-Any ChrF on 30 new languages. The x-axis depicts the amount of monolingual data available for the language, while the y-axis depicts the ChrF score of the 1.6B parameter Transformer model after fine-tuning with online back-translation. The data point corresponding to each language is represented by its language code.}
  \label{fig:lang200chrf}
\end{figure}

For xx$\rightarrow$en translation, translation quality is not well correlated with the amount of monolingual data available for the language. Languages which perform well are typically ones that have similar languages in the supervised set (South Asian languages, Central Asian languages, Pidgins and Creoles), agreeing with work on multilingual natural language understanding that highlights the role of related language supervision for downstream zero-resource cross-lingual tasks \citep{Lauscher2020FromZT,Turc2021RevisitingTP}. On the other hand, en$\rightarrow$xx translation BLEU is high only for languages which have high xx$\rightarrow$en translation quality, and are on the right half of the plot (relatively large amounts of monolingual data). We also plot the ChrF scores for en$\rightarrow$xx translation with these models in Figure \ref{fig:lang200chrf}. We find that the relative performance of agglutinative and poly-synthetic languages (for example, Oromo (om) and Ganda (lg)) is under-represented when reporting BLEU.

\section{Conclusion}
One of the biggest challenges en route to achieving universal translation between any human language pair is large-scale training data collection. Due to the prohibitive cost of curating parallel data for all language pairs, we have to adopt a practical stance; our training procedure should be able to leverage any available type of data and training supervision. In this work we demonstrate that by mixing supervised and self-supervised learning techniques, a multilingual model can learn to translate effectively even for severely under-resourced language pairs with no parallel data and little monolingual data, approaching or passing the quality of models trained on supervised parallel data. We evaluate our procedure on WMT as well as on an in-house, web-mined dataset covering over 206 languages, and demonstrate the feasibility of scaling up to hundreds of languages without the need for parallel data annotations. Observations from our experiments answer multiple research questions regarding the cross-lingual transfer at a large scale. We hope that the work presented here lays a milestone towards reaching the goal of universal translation between thousands of languages.

\bibliography{tacl2018}
\bibliographystyle{acl_natbib}

\clearpage

\appendix

\section{Details on which languages have similar embeddings}
\label{appendix:linguistic}

\begin{table*}[]
\centering
\begin{tabular}{l|llll}
\toprule
group/lang          & ro                 & gu              & kk           & et \\
\midrule
4 related & \textbf{es\textsuperscript{†},cs\textsuperscript{†},ru} & \textbf{hi},zh,tr,kk    &  hi,zh,\textbf{tr},gu &  \textbf{fi\textsuperscript{†}},lt\textsuperscript{†},lv\textsuperscript{†},de\textsuperscript{†}  \\
2 related & \textbf{fr\textsuperscript{†},cs\textsuperscript{†}},et\textsuperscript{†},\textbf{hi}  & \textbf{hi},zh,\textbf{fr,cs}  & hi,zh,fr,cs  &  \textbf{fi\textsuperscript{†}},lt\textsuperscript{†},hi,zh  \\
0 related & fi\textsuperscript{†},zh,et\textsuperscript{†},\textbf{hi}     & \textbf{es,ru,fr,cs} & es,ru\textsuperscript{†},fr,cs & ru,es\textsuperscript{†},hi,zh \\
\bottomrule
\end{tabular}
\caption{The groupings of ``similar" languages corresponding to Table \ref{tab:similarity}. Languages in \textbf{bold} are in the same language family, and those with daggers (†) share a script.}
\label{appendix:tab:similarity}
\end{table*}

In Section \ref{main:linguistic}, we discussed results of including languages that are ``similar" according to the chart presented by \citet{kudugunta2019investigating}, even if in some cases they are not categorized as similar by linguists. Table \ref{appendix:tab:similarity} shows the groupings we used, marked by whether they share a language family (Indo-European, Turkic, Uralic) or a script (Latin, Gujarati, or Cyrillic). With the exception of Romanian and to some extent Estonian, these traditional measures of similarity do not relate well to our groupings.

For Romanian, the picture is clear: adding more Indo-European languages helps, and adding non-Indo-European languages hurts. However, for Gujarati, the most ``similar" languages are mostly non-Indo-European, and all of the ``dissimilar" languages in fact \textit{are} Indo-European languages. One way of looking at the groupings is that it seems like there are three categories of languages from the model's perspective: European Indo-European languages (aka Italic + Slavic), Indo-Aryan languages, and everything else.

\section{Choice of languages in the Analysis section}
For Table \ref{tab:numparalleldirections}, we choose four languages with different scripts and different resource settings to present a complete picture. In Table \ref{tab:numunsuperviseddirections}, we choose the five languages with lowest amount of data in order to have all the results by running fewer experiments. Choosing to evaluate on these five language pairs helped us get all the results training only one model for rows 2, 3 and 4. For Table \ref{tab:similarity}, the choice of languages was influenced by which languages have similarities among the pool of fifteen available languages. In Table \ref{tab:amount}, the amount of supervised and unsupervised data available influenced the pick. We wanted to have enough parallel data in the supervised training pool, so we included fr, es, cs, and ru for all experiments, meaning that we could not evaluate on these languages. In Table \ref{tab:domain}, lt was the only language with sufficient monolingual and parallel data in different domains. Finally, for Table \ref{tab:stacking}, we report results on 2 language pairs for which we had the most significant changes from the way these methods were stacked during training.

\begin{table*}[t]
\centering
\begin{tabular}{l|cc|cc|cc|cc|cc}
\toprule
                   & \multicolumn{2}{c|}{En-Gu}    & \multicolumn{2}{c|}{En-Tr}     & \multicolumn{2}{c|}{En-Kk}     & \multicolumn{2}{c|}{En-Hi}    & \multicolumn{2}{c}{En-Ro}    \\
Para Size          & \multicolumn{2}{c|}{0.01m}    & \multicolumn{2}{c|}{0.2m}      & \multicolumn{2}{c|}{0.2m}      & \multicolumn{2}{c|}{0.3m}     & \multicolumn{2}{c}{0.6m}     \\
Mono Size          & \multicolumn{2}{c|}{4.6m}     & \multicolumn{2}{c|}{9.6m}      & \multicolumn{2}{c|}{13.8m}     & \multicolumn{2}{c|}{23.6m}    & \multicolumn{2}{c}{14.1m}    \\

            Direction       & $\leftarrow$ & $\rightarrow$ & $\leftarrow$  & $\rightarrow$ & $\leftarrow$  & $\rightarrow$ & $\leftarrow$ & $\rightarrow$ & $\leftarrow$ & $\rightarrow$ \\
                   \midrule
Multilingual & 1.0          & 0.6           & \textbf{15.9}          & 13.6          & 11.5          & 1.9           & 8.5          & 4.5           & 30.1         & 23.4          \\
Zero-resource               & \textbf{14.8}         & \textbf{11.7}          & 15.8          & \textbf{15.4}          & \textbf{12.8}          & \textbf{9.3}           & \textbf{17.2}         & \textbf{11.3}          & \textbf{36.8}         & \textbf{25.4}          \\
\midrule
\midrule
                   & \multicolumn{2}{c|}{En-Lt}    & \multicolumn{2}{c|}{En-Lv}     & \multicolumn{2}{c|}{En-Et}     & \multicolumn{2}{c|}{En-De}    & \multicolumn{2}{c}{En-Fi}    \\
Para Size          & \multicolumn{2}{c|}{0.6m}     & \multicolumn{2}{c|}{0.6m}      & \multicolumn{2}{c|}{2.1m}      & \multicolumn{2}{c|}{4.5m}     & \multicolumn{2}{c}{6.5m}     \\
Mono Size          & \multicolumn{2}{c|}{106.1m}   & \multicolumn{2}{c|}{10.2m}     & \multicolumn{2}{c|}{51.6m}     & \multicolumn{2}{c|}{275m}     & \multicolumn{2}{c}{18.8m}    \\
      Direction             & $\leftarrow$ & $\rightarrow$ & $\leftarrow$  & $\rightarrow$ & $\leftarrow$  & $\rightarrow$ & $\leftarrow$ & $\rightarrow$ & $\leftarrow$ & $\rightarrow$ \\
      \midrule
Multilingual & 21.3         & \textbf{11.5}          & 15.0          & 14.2          & \textbf{23.1} & 18.2          & 31.7         & \textbf{29.9}          & \textbf{27.3}         & \textbf{18.1}          \\
Zero-resource           & \textbf{24.4}         & 8.9           & \textbf{20.9} & \textbf{15.1} & 22.1          & \textbf{18.9} & \textbf{35.9}         & 26.7          & 23.8         & 17.8          \\
\midrule
\midrule
                   & \multicolumn{2}{c|}{En-Es}    & \multicolumn{2}{c|}{En-Zh}     & \multicolumn{2}{c|}{En-Ru}     & \multicolumn{2}{c|}{En-Fr}    & \multicolumn{2}{c}{En-Cs}    \\
Para Size          & \multicolumn{2}{c|}{15.1m}    & \multicolumn{2}{c|}{25.9m}     & \multicolumn{2}{c|}{38.4m}     & \multicolumn{2}{c|}{40.4m}    & \multicolumn{2}{c}{64.3m}    \\
Mono Size          & \multicolumn{2}{c|}{43.8m}    & \multicolumn{2}{c|}{2.1m}      & \multicolumn{2}{c|}{80.1m}     & \multicolumn{2}{c|}{160.9m}   & \multicolumn{2}{c}{72.1m}    \\
         Direction          & $\leftarrow$ & $\rightarrow$ & $\leftarrow$  & $\rightarrow$ & $\leftarrow$  & $\rightarrow$ & $\leftarrow$ & $\rightarrow$ & $\leftarrow$ & $\rightarrow$ \\
         \midrule

Multilingual & \textbf{32.7}         & \textbf{31.1}          & \textbf{21.7}          & \textbf{31.3}          & \textbf{36.0}          & \textbf{26.4}          & \textbf{37.2}         & \textbf{41.3}          & \textbf{31.3}         & \textbf{23.8}          \\
Zero-resource           & 29.1         & 28.5          & 16.9          & 21.2          & 30.6          & 25.5          & 36.1         & 33.2          & 28.3         & 22.1 \\
\bottomrule
\end{tabular}
\caption{Comparing the results of our zero-resource models, against bilingual baselines of same capacity. The languages have been sorted from low to high resource depending on the availability of the parallel data.} \label{appendix:tab:bilingual}
\end{table*}

\begin{table*}[h]
\centering
\label{table:dataset_bi}
\begin{tabular}{ccccrrr}
\toprule
 \multirow{2}{1.5cm}{\centering Language Pair} & \multicolumn{3}{c}{Data Sources} & \multicolumn{3}{c}{$\#$ Samples}\\
 \cmidrule{2-7}
  & Train & Dev & Test & Train & Dev & Test\\
\midrule
cs$\rightarrow$en                                                            & WMT'19    & WMT'17    & WMT'18   & 64336053     & 3005    & 2983    \\
fr$\rightarrow$en                                                            & WMT'15    & WMT'13    & WMT'14   & 40449146     & 3000    & 3003    \\
ru$\rightarrow$en                                                            & WMT'19    & WMT'18    & WMT'19   & 38492126     & 3000    & 2000    \\
zh$\rightarrow$en                                                            & WMT'19    & WMT'18    & WMT'19   & 25986436     & 3981    & 2000    \\
es$\rightarrow$en                                                            & WMT'13    & WMT'13    & WMT'13   & 15182374     & 3004    & 3000    \\
fi$\rightarrow$en                                                            & WMT'19    & WMT'18    & WMT'19   & 6587448      & 3000    & 1996    \\
de$\rightarrow$en                                                            & WMT'14    & WMT'13    & WMT'14   & 4508785      & 3000    & 3003    \\
et$\rightarrow$en                                                            & WMT'18    & WMT'18    & WMT'18   & 2175873      & 2000    & 2000    \\
lv$\rightarrow$en                                                            & WMT'17    & WMT'17    & WMT'17   & 637599       & 2003    & 2001    \\
lt$\rightarrow$en                                                            & WMT'19    & WMT'19    & WMT'19   & 635146       & 2000    & 1000    \\
ro$\rightarrow$en                                                            & WMT'16    & WMT'16    & WMT'16   & 610320       & 1999    & 1999    \\
hi$\rightarrow$en                                                            & WMT'14    & WMT'14    & WMT'14   & 313748       & 520     & 2507    \\
kk$\rightarrow$en                                                            & WMT'19    & WMT'19    & WMT'19   & 222424       & 2066    & 1000    \\
tr$\rightarrow$en                                                            & WMT'18    & WMT'17    & WMT'18   & 205756       & 3007    & 3000    \\
gu$\rightarrow$en                                                            & WMT'19    & WMT'19    & WMT'19   & 11670        & 1998    & 1016    \\
\midrule
en$\rightarrow$cs                                                            & WMT'19    & WMT'17    & WMT'18   & 64336053     & 3005    & 2983    \\
en$\rightarrow$fr                                                            & WMT'15    & WMT'13    & WMT'14   & 40449146     & 3000    & 3003    \\
en$\rightarrow$ru                                                            & WMT'19    & WMT'18    & WMT'19   & 38492126     & 3000    & 2000    \\
en$\rightarrow$zh                                                            & WMT'19    & WMT'18    & WMT'19   & 25986436     & 3981    & 2000    \\
en$\rightarrow$es                                                            & WMT'13    & WMT'13    & WMT'13   & 15182374     & 3004    & 3000    \\
en$\rightarrow$fi                                                            & WMT'19    & WMT'18    & WMT'19   & 6587448      & 3000    & 1996    \\
en$\rightarrow$de                                                            & WMT'14    & WMT'13    & WMT'14   & 4508785      & 3000    & 3003    \\
en$\rightarrow$et                                                            & WMT'18    & WMT'18    & WMT'18   & 2175873      & 2000    & 2000    \\
en$\rightarrow$lv                                                            & WMT'17    & WMT'17    & WMT'17   & 637599       & 2003    & 2001    \\
en$\rightarrow$lt                                                            & WMT'19    & WMT'19    & WMT'19   & 635146       & 2000    & 1000    \\
en$\rightarrow$ro                                                            & WMT'16    & WMT'16    & WMT'16   & 610320       & 1999    & 1999    \\
en$\rightarrow$hi                                                            & WMT'14    & WMT'14    & WMT'14   & 313748       & 520     & 2507    \\
en$\rightarrow$kk                                                            & WMT'19    & WMT'19    & WMT'19   & 222424       & 2066    & 1000    \\
en$\rightarrow$tr                                                            & WMT'18    & WMT'17    & WMT'18   & 205756       & 3007    & 3000    \\
en$\rightarrow$gu                                                            & WMT'19    & WMT'19    & WMT'19   & 11670        & 1998    & 1016    \\
\bottomrule
\end{tabular} 
\caption{Data sources and number of samples for the parallel data in our corpus.} \label{appendix:tab:paralleldata}
\end{table*}

\appendix

\begin{table*}[h]
\centering
\label{table:datasets_mono}
\resizebox{\linewidth}{!}{
\begin{tabular}{lccccccrrr}
\toprule
\multirow{2}{1.4cm}{\centering Language} & \multicolumn{6}{c}{\centering Data Sources} & \multicolumn{3}{c}{$\#$ Samples}\\
\cmidrule{2-10}
& News Crawl & News Commentary & Common Crawl & Europarl & News Discussions & Wiki Dumps & Train & Dev & Test\\
\midrule
en & \checkmark & & & & & & 199900557 & 3000 & 3000\\
ro & \checkmark & & & & & & 14067879 & 3000 & 3000\\
de & \checkmark & & & & & & 275690481 & 3000 & 3000\\
fr & \checkmark & \checkmark & & \checkmark & \checkmark & & 160933435 & 3000 & 3000\\
cs & \checkmark & & & & & & 72157988 & 3000 & 3000\\
es & \checkmark & & & & & & 43814290 & 3000 & 3000\\
et & \checkmark & & \checkmark & & & & 51683012 & 3000 & 3000\\
fi & \checkmark & & & \checkmark & & & 18847600 & 3000 & 3000\\
gu & \checkmark & & \checkmark & & & & 4644638 & 3000 & 3000\\
hi & \checkmark & & & & & & 23611899 & 3000 & 3000\\
kk & \checkmark & \checkmark & \checkmark & & & \checkmark & 13825470 & 3000 & 3000\\
lt & \checkmark & & \checkmark & \checkmark & & \checkmark & 106198239 & 3000 & 3000\\
lv & \checkmark & & & \checkmark & & & 10205015 & 3000 & 3000\\
ru & \checkmark & & & & & & 80148714 & 3000 & 3000\\
tr & \checkmark & & & & & & 9655009 & 3000 & 3000\\
zh & \checkmark & \checkmark & & & & & 2158309 & 3000 & 3000\\
\bottomrule
\end{tabular}} 
\caption{Data sources and number of samples for the monolingual data in our corpus.} \label{appendix:tab:monolingualdata}
\end{table*}

\end{document}